\documentclass[10pt,twocolumn,letterpaper]{article}

\usepackage{3dv}
\usepackage{times}
\usepackage{epsfig}
\usepackage{graphicx}
\usepackage{amsmath}
\usepackage{amssymb}

\usepackage[pagebackref=false,breaklinks=true,colorlinks,bookmarks=false]{hyperref}

\threedvfinalcopy %

\def\threedvPaperID{428} %

\ifthreedvfinal\fi

\usepackage{graphics} %
\usepackage{mathptmx} %
\usepackage{xcolor}
\usepackage{appendix}
\usepackage{subcaption}
\usepackage[normalem]{ulem} %

\usepackage{blindtext}
\usepackage[
singlelinecheck=false %
]{caption}

\graphicspath{{./figs/}}

\newcommand{\GS}[1]{#1}

\newcommand{\m}{\rm m}

\renewcommand{\log}{\rm log} %

\usepackage{calrsfs}
\DeclareMathAlphabet{\pazocal}{OMS}{zplm}{m}{n}
\newcommand{\Loss}{\pazocal{L}}

\newcommand{\MethodName}{\textcolor{black}{DANCE}}
\newcommand{\MethodNameLong}{\textcolor{black}{Domain Adaptation of Networks for Camera pose Estimation}}

\begin{document}
\title{Domain Adaptation of Networks for Camera Pose Estimation: \\
Learning Camera Pose Estimation Without Pose Labels}
\author{Jack Langerman$^{1}$, Ziming  Qiu$^{1,3}$, G\'abor S\"or\"os$^{2}$, D\'avid Seb\H{o}k$^{2}$, Yao Wang$^{3}$, Howard Huang$^{1}$%
\thanks{JL and ZQ contributed equally. This work was carried out while ZQ and DS interned at Nokia Bell Labs in 2020. Corresponding author: \tt\small howard.huang@nokia-bell-labs.com\newline
$^{1}$Nokia Bell Labs, Murray Hill, NJ, USA\newline
$^{2}$Nokia Bell Labs, Budapest, Hungary\newline
$^{3}$New York University, New York, NY, USA\newline
}
}

\maketitle

\begin{abstract}
One of the key criticisms of deep learning is that large amounts of expensive and difficult-to-acquire training data are required in order to train models with high performance and good generalization capabilities. Focusing on the task of monocular camera pose estimation via scene coordinate regression (SCR), we describe a novel method, \MethodNameLong\, (\MethodName), which enables the training of models without access to any labels on the target task. \MethodName\, requires unlabeled images (without known poses, ordering, or scene coordinate labels) and a 3D representation of the space (e.g., a scanned point cloud), both of which can be captured with minimal effort using off-the-shelf commodity hardware. \MethodName\, renders labeled synthetic images from the 3D model, and bridges the inevitable domain gap between synthetic and real images by applying unsupervised image-level domain adaptation techniques (unpaired image-to-image translation). When tested on real images, the SCR model trained with \MethodName\, achieved comparable performance to its fully supervised counterpart (in both cases using PnP-RANSAC for final pose estimation) at a fraction of the cost. Our code and dataset are available at \url{https://github.com/JackLangerman/dance}
\end{abstract}

\section{INTRODUCTION}
Estimating the 3D position and 3D orientation (6DoF pose) of an agent or an object with respect to a reference coordinate frame is a fundamental requirement in robotics applications. Visual localization offers several advantages compared to other modalities for deriving 6DoF poses: it is effective both indoors and outdoors, it requires no extra infrastructure, and it can be precise and accurate using only a single RGB image.

\begin{figure}[ht!]
\centerline{\includegraphics[width=\columnwidth]{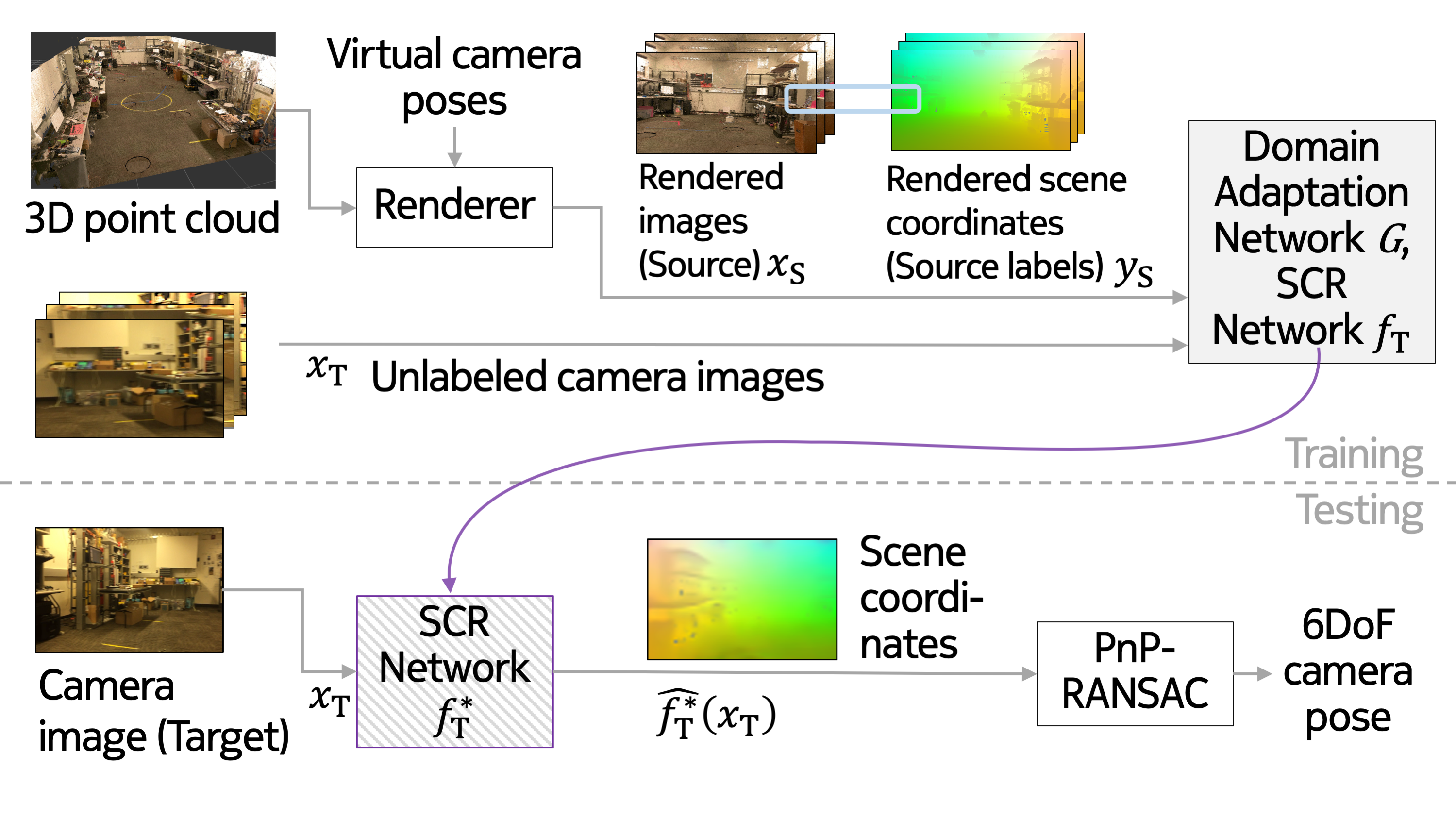}}
\caption{At training time, the SCR network is trained using unsupervised deep domain adaptation techniques, which bridge the domain gap between labeled synthetic images and real camera images of the scene. At test time, only the trained SCR network is kept, it regresses 3D scene coordinates for each pixel, from which the camera pose is calculated via PnP-RANSAC.}
\label{overall_pipeline}
\vspace{-0.5cm}
\end{figure}

Prior solutions for pose estimation from RGB images can be split into two categories: those that use hand-crafted features and those based on machine learning. The methods based on hand-crafted features incorporate extensive prior knowledge about the problem, and often achieve better accuracy today. However, learned methods can offer higher speed, robustness to occlusions, and access to continuous intermediate representations, and are therefore gaining popularity. Furthermore, learning-based methods such as Convolutional Neural Networks (CNNs) can offer potentially more compact representations of spaces in their weights compared to Simultaneous Localization and Mapping (SLAM) maps, and have been shown to outperform many hand-crafted methods in textureless areas~\cite{Walch2017PoseLSTM}.

The major drawback of learning-based methods in the past has been that they require supervised training on large image sets with known camera poses, and need to be re-trained for every new scene or in case any changes in the target scene.
In general, a labeled training set needs to be generated for each new environment, making data collection tedious and hindering the scalability and practical applicability of these techniques. A typical way to generate a training set is to track the camera with an external localization system while it moves through the environment, implying high overhead. Infrared tracking systems using fixed infrastructure cameras and active beacons (e.g., WorldViz or Vicon) can capture sub-cm and sub-degree accurate pose labels for the moving camera images. However, the cost and difficulty of using these localization systems is not trivial and is even infeasible in some places. %
Another method of obtaining labeled training images is to use on-board RGB-D SLAM as in 7Scenes~\cite{Shotton2013SCR} and ScanNet~\cite{Dai2017ScanNet}. 

\GS{In this paper, we take a representative method that solves the scene coordinate regression problem, and show that it can be trained with synthetically generated images at a fraction of the cost compared to acquiring real pose labels, and it still achieves the same median error.}
As an alternative to training with expensive labeled camera images, we propose a novel pipeline \MethodNameLong\, (\MethodName), shown in Fig.~\ref{overall_pipeline}, which relies on a lower-cost combination of unlabeled camera images and labeled synthetic images. 
Our key contributions are (i) providing a domain adaptation methodology to train a neural network for the task of scene coordinate regression (SCR) using only labels generated in simulation (via rendering). This enables (ii) the training of camera pose estimation at significantly lower overhead cost compared to fully supervised learning-based solutions.
Specifically, we render a large number of images with known (arbitrary) poses and scene coordinates from a laser scan of the space. The appearance of these images is far from real photos, and to bridge this domain gap, (iii) we employ domain adaptation at the image level. These techniques allow the training of the SCR network with domain adapted labeled rendered images only. We also (iv) release our collected dataset with training and testing code\footnote{\url{https://github.com/JackLangerman/dance}}. %

\section{RELATED WORK}

Vision-based localization approaches can be categorized into retrieval-based and regression-based families, both requiring a large number of images that cover the whole scene.

Retrieval-based methods typically extract  global descriptors from keyframes \cite{Cummins2011FabMap2} \cite{Vysotska2019}
and/or local descriptors from feature points
\cite{Straub2013Binary} \cite{MurArtal2017ORBSLAM2} \cite{MunozSalinas2020UcoSLAM},
and build a database of scene descriptors. Then, descriptors extracted from the query image or sequence of images are matched to the closest entries in the database and assigned a location, which is finally validated by geometric constraints. Feature point-based methods are more robust, but tend to be slower than keyframe-based methods.
While we focus on indoor scenarios, our problem is highly related to large-scale visual localization methods \cite{Sattler2017PAMI} \cite{Humenberger2020Kapture} \cite{Sarlin2019HLoc} \cite{Lynen2020} which are predominantly based on features and sparse 3D maps, but have recently begun incorporating learning-based components as well~\cite{Sarlin21pixloc}. 
Prior knowledge about the coarse location from
GPS \cite{Sattler2017PAMI} \cite{Zeisl2015} \cite{Lynen2020}, 
radio signals \cite{Ishihara2018RadioVisual} \cite{Hashemifar2019}, 
a LiDAR map~\cite{Dube2019SegMap}
or other means can significantly reduce the search space and can make these methods applicable even at city scale~\cite{Lynen2020}.

Regression-based methods are both robust and fast, and therefore offer a promising new direction. However, at the time of writing, they are less accurate, limited in scale, and expensive to train.
One family of regression approaches use learned models \cite{Shotton2013SCR} \cite{Valentin2015} \cite{Brachmann2016} \cite{Cavallari2017} to perform SCR and then input point samples from the intermediate scene coordinate map to a PnP-RANSAC pose estimator.
More recent examples of this group of methods are DSAC, DSAC++, and DSAC\text{*}~\cite{Brachmann2020DSACStar}, which add differentiable approximations for all steps of the pipeline, including SCR, PnP, and RANSAC and achieve state-of-the-art pose estimation performance.
Another family of regression approaches including PoseNet~\cite{Kendall2015PoseNet}, PoseLSTM~\cite{Walch2017PoseLSTM}, and RelocNet~\cite{Balntas2018RelocNet} directly return the 6DoF pose from a single image and can be trained end to end.
In addition to localization of a single image, VLocNet++~\cite{Radwan2018VLocNetPP} implements learning-based odometry and adds semantics, and for the first time exceeds the accuracy of feature-based methods.
\GS{Sattler et al.~\cite{Sattler2019} analyzed why direct pose regression methods generally fall behind feature-based methods and concluded that CNNs rather learn to retrieve similar images instead of learning a 3D map of the space. Further research is urged for, and our training technique makes that a lot easier than before.}

An often cited drawback of learning-based methods is that they are trained for a particular scene and are difficult to adapt to other environments. New techniques have attempted to address this drawback using a variety of methods. The authors of \cite{Cavallari2017} and \cite{Cavallari2020} show how to adapt a random forest to a new place at runtime. RelocNet~\cite{Balntas2018RelocNet} performs regression of relative poses and thus avoids the need to retrain for every scene, while ESAC~\cite{Brachmann2019ESAC} breaks a scene into smaller parts and trains a network for each before using an ensemble network to decide which subnetwork to use, thus allowing SCR to be performed for larger areas. 
Unfortunately, all regression models
require supervised training on image datasets with ground truth pose labels or scene coordinates, which are very expensive to acquire. 
Similarly in feature-based methods, the need to support the innumerable variations in scenes resulting from lighting changes, weather conditions, etc., and the cost of collecting data across these conditions can be prohibitively expensive. These challenges can be avoided by synthetically generating rather than collecting data \cite{Mashita2016} \cite{Mueller2019} \cite{Sibbing2013PCD2SIFT} \cite{Taira2021InLoc} \cite{Torii2015ViewSynthesis}.

Because modeling every aspect of the real world in a rendering pipeline is infeasible, synthetic images inevitably differ from those captured with a real camera -- this difference in appearance is referred to as the \emph{domain gap}.
The authors of \cite{Shoman2018RESTarxiv} demonstrate that feature-based retrieval using the representations learned by a PoseNet trained on purely synthetic images is highly effective for synthetic queries, but fails when used on real images.

Researchers have taken a variety of approaches to attempt to reduce this domain gap. \cite{Shoman2018RESTarxiv} transforms real features to look more similar to synthetic features using an autoencoder.
Other recent works on 6DoF object pose estimation~\cite{Akkaya2019Rubik}~\cite{Li20206DoF}~\cite{Wen2020Se3TrackNet} apply domain randomization, i.e.,  generating synthetic training data with randomized rendering parameters in order to robustify the trained networks.
Several domain adaptation works~\cite{ganin2015unsupervised}~\cite{tzeng2017adversarial} apply feature-level alignment for image classification; other recent works \cite{Hoffman2018CyCADA}~\cite{li2019bidirectional} combine the CycleGAN-based image-level alignment, and adversarial feature-level alignment for image segmentation. CyCADA~\cite{Hoffman2018CyCADA} performs both the image-level and feature-level adaptation in an end-to-end manner while BDL~\cite{li2019bidirectional} decouples them. 
Because both image segmentation and our SCR task require dense predictions, it is imperative for the adapted synthetic images to preserve both the semantic content and the geometric structure. We propose to employ the contrastive unpaired translation (CUT) model \cite{park2020contrastive} for image-level domain adaptation to train the SCR network.

\section{METHOD}
\GS{In order to successfully bridge the domain gap, we must transform rendered images such that they appear to come from the same distribution as the real camera images while preserving both the semantic content and the geometric structure to enable effective camera pose estimation.}
We utilize a generative adversarial network (GAN) based framework (the CUT model \cite{park2020contrastive}) for the image-to-image translation step. Specifically, using the rendered images $X_S$ along with a set of unordered, unlabeled photos $X_T$ from the same scene, a mapping network $G_{S \rightarrow T}$ is trained using the CUT framework to map from the source domain of rendered images $X_S$ to the target domain of real photos $X_T$ without changing the geometric and semantic content in $X_S$. In this way, the corresponding rendered scene coordinate labels $Y_S$ of $X_S$ can be reused for $\hat{X}_T$, where $\hat{X}_T$ = $G^*_{S \rightarrow T}(X_S)$ ($^*$ denotes converged models after training). Finally, an SCR network $f_T$ is trained with ($\hat{X}_T$, $Y_S$) for use in the target domain $X_T$. Note that direct training on domain $X_T$ is not possible because the target labels $Y_T$ are not available and real photos $X_T$ do not have any corresponding or pairwise relationship with synthetic images $X_S$. At inference time, we apply the target network $f_T^*$ in the target domain $X_T$ (real photos) and feed the predicted scene coordinates $f_T^*(X_T)$ to a traditional PnP-RANSAC~\cite{ransac81} to compute the final pose estimates. This whole process is illustrated in Figure~\ref{overall_pipeline}. 

One could ask why not just train on ($X_S$, $Y_S$) and apply the converged model in the target domain $X_T$ directly. As shown in Table \ref{table:1} (a) (blind transfer), this approach leads to severely degraded performance. In order to bridge the domain gap 
between the rendered images $X_S$ and photos from the real world $X_T$, we train the SCR network on the domain adapted labeled rendered images ($\hat{X}_T$, $Y_S$).

\subsection{Mathematical Description}

We consider an unsupervised domain adaptation problem, where we are provided distributions for source data $X_S$, source labels $Y_S$ and target data $X_T$, but no target labels. The ultimate goal is to learn a model $f_T$ that can accurately predict the label on the target distribution $X_T$. In our problem, $X_S$ indicates rendered (source domain) images, $Y_S$ indicates rendered (source domain) scene coordinate labels, $X_T$ indicates real (target domain) camera images, and $f_T$ indicates the SCR network trained using \MethodName\ to perform well in the domain of real camera images (target domain).
The rendered scene coordinates $Y_S$ encode the $(X,Y,Z)$ coordinates in the model of the world (point cloud) which correspond to each pixel $(U,V)$ in the images $X_S$. The objective of the trained SCR network $f_T$ is to predict these $((U,V),(X,Y,Z))$ correspondences for images in the target domain. The training procedure is illustrated in Fig.~\ref{fig:training_pipeline} and described below.

\subsubsection{Domain Adaptation}
We use a combination of pre-processing and image-level domain adaptation techniques to map the source images into the target domain. First, simple histogram matching brings the color distribution of the rendered images $X_S$ closer to that of the target domain $X_T$, \GS{because our laser scanner's automatic white balance setting leads to a mismatch with the query camera.} Hereafter, $X_S$ indicates source rendered images after histogram matching. 

\begin{figure}[!t]
\vspace{0.25cm}
\includegraphics[width=1.0\columnwidth]{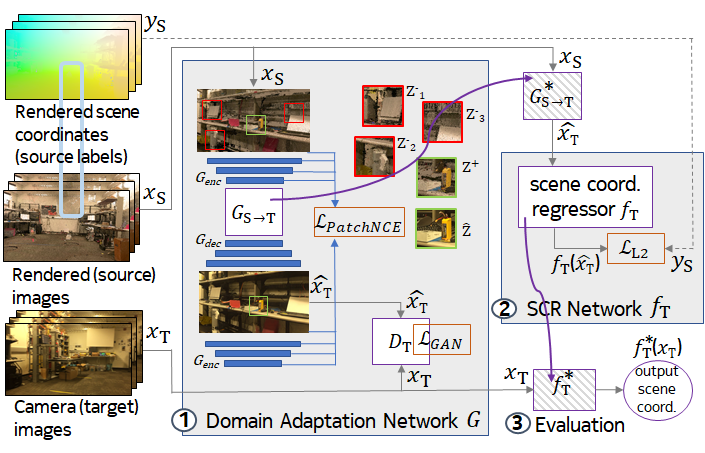}
\caption{The training pipeline for the SCR network with unsupervised deep domain adaptation.}
\vspace{-0.5cm} %
\label{fig:training_pipeline}
\end{figure}

Next, an image-level domain adaptation network $G_{S \rightarrow T}$ is trained to translate the rendered images (source) $X_S$ into the domain of real photos (target) $X_T$ so that they can fool an adversarial discriminator network $D_{T}$  (see Fig.~\ref{fig:training_pipeline}). 
The following GAN objective is employed:
\begin{equation} 
\label{eq2}
\begin{split}
&\Loss_{GAN}(X_S, X_T; G_{S \rightarrow T}, D_{T}) \\ = & \mathbb{E}_{x_t \sim X_T} \left [ \log D_{T}(x_t) \right ] \\ + & \mathbb{E}_{x_s \sim X_S} \left [ \log(1 - D_{T}(G_{S \rightarrow T}(x_s))) \right ]
\end{split}
\end{equation}
The mapping network $G_{S \rightarrow T}$ tries to minimize the loss function while the discriminator $D_{T}$ tries to maximize it. This optimization procedure ensures that the learned mapping $G_{S \rightarrow T}$ is able to translate source images to convincing target images. Note that there is no corresponding or pairwise relationship between source data $X_S$ and target data $X_T$.

Because scene coordinate estimation from the input image is a dense prediction task, it is important that the mapping $G_{S \rightarrow T}(X_S)$ preserves the structure and content of the original image $X_S$. However, a traditional GAN model (with objective \ref{eq2} only) can not ensure this consistency requirement. To enforce consistency, the following multi-layer patch-wise noise-contrastive estimation (PatchNCE) loss is adopted \cite{park2020contrastive}:
\vspace{-0.1cm} %
\begin{equation} 
\label{eq3}
\begin{gathered}
\Loss_{PatchNCE}(X_S; G_{S \rightarrow T}, H) =\\ \mathbb{E}_{x_s \sim X_S} \sum_{l=1}^{L}\sum_{s=1}^{S} \l(\hat{z}_{l}^{(s)}, z_{l}^{+(s)}, z_{l}^{-(S \setminus s)})
\end{gathered}
\end{equation}
A high-level understanding of the $\Loss_{PatchNCE}$ loss \cite{park2020contrastive} is that the patches at the same location before and after the mapping network should be more similar than patches at other spatial locations in the same image. Here, $\hat{z}_{l}^{(s)}$ is the feature vector at location $s$ from the $l$-th feature layer for the translated image $\hat{x}_{T}$ = $G_{S \rightarrow T}(x_{S})$. $ z_l^{+(s)} $ is the feature vector at the same location $s$ from the $l$-th layer for the input image $x_S$, and $z_{l}^{-(S \setminus s)}$ are the feature vectors at locations other than $s$ from the $l$-th layer for the input image $x_S$.

The mapping network $G_{S \rightarrow T}$ can be decomposed into an encoder $G_{enc}$ followed by a decoder $G_{dec}$. The feature vectors $\hat{z}_{l}^{(s)}$, $z_{l}^{+(s)}$ and $z_{l}^{-(S \setminus s)}$ are extracted from the $l$-th layer of $G_{enc}$ and then passed through a small 2-layer multi-layer perceptron (MLP) network $H_{l}$, producing 256-dim final features. $L$ (=5) is the number of layers chosen to extract features and $S$ (=256) is the number of locations sampled in each layer. Each layer and spatial location of these extracted features represents a patch of the input image, with deeper layers corresponding to larger patches. The loss function
\begin{equation}
    \begin{gathered}
    \l(\hat{z}_{l}^{(s)}, z_{l}^{+(s)}, z_{l}^{-(S \setminus s)}) = \\
    -\log\left [ \frac{exp(\hat{z}_{l}^{(s)}\cdot  z_{l}^{+(s)}/\tau)}{exp(\hat{z}_{l}^{(s)}\cdot  z_{l}^{+(s)}/\tau) + \sum_{n=1}^{len(S \setminus s)}exp(\hat{z}_{l}^{(s)}\cdot z_{l}^{-(n)}/\tau)} \right ]
    \end{gathered}
\end{equation}
encourages the current query $\hat{z}_{l}^{(s)}$ to be closer to the positive example $z_{l}^{+(s)}$ but different from negatives $z_{l}^{-(S \setminus s)}$. $\tau$ (=0.07) is a temperature parameter to scale the similarity between two examples. For more details, please refer to \cite{park2020contrastive}.

The aggregate loss used to train the image level adaptation network $G_{S \rightarrow T}$ can be summarized as follows:
\begin{equation} \label{eq5}
\begin{split}
&\Loss_{CUT}(X_S, X_T; G_{S \rightarrow T}, D_{T}, H) \\&= \lambda_{GAN} \Loss_{GAN}(X_S, X_T; G_{S \rightarrow T}, D_{T}) \\&+ \lambda_{S} \Loss_{PatchNCE}(X_S; G_{S \rightarrow T}, H) \\&+ \lambda_{T}\Loss_{PatchNCE}(X_T; G_{S \rightarrow T}, H)
\end{split}
\end{equation}
The third term $\Loss_{PatchNCE}(X_T; G_{S \rightarrow T}, H)$ is an identity loss for network regularization.

After training the domain adaptation network, we only keep the trained model $G^*_{S \rightarrow T}$ as a fixed transformation function from source images $X_S$ (rendered images) to target images $X_T$. Hereafter, $\hat{X}_T$ = $G^*_{S \rightarrow T}(X_S)$ indicates domain adapted images.

\subsubsection{Target SCR Network Training}

Finally, the paired training data ($\hat{X}_T$, $Y_S$) are used to train a target SCR model $f_T$ with an L2 loss:
\begin{equation} \label{eq4}
\Loss_{L2}(\hat{X}_T, Y_S; f_T) = \mathbb{E}_{(\hat{x}_t,y_s)\sim(\hat{X}_T,Y_S)} \left \| y_s-f_T(\hat{x}_t) \right \|_{2}
\end{equation}
After training the target model $f_T$ with the above loss, the trained model $f^*_T$ can be used to predict scene coordinates for target camera images $X_T$ at testing time. Finally, the predicted coordinates $f^*_T(X_T)$ are passed to PnP-RANSAC to compute final pose estimates.

\subsection{Implementation Details}

\subsubsection{Network Architectures}
The SCR network $f_T$ is a fully convolutional network consisting of a feature encoder followed by a regression head. Specifically, the feature encoder is a ResNet18 \cite{HeResNet2015} after removing the last 2 layers (1000-d fc and average pool) and setting the last 2 stride-2 convolutional layers (conv4\_1 and conv5\_1) to stride 1. The regression head has 3 convolutional layers to transform the features from the encoder to the 3-channel scene coordinate predictions. For the image-level adaptation, we follow the network architectures of CUT \cite{park2020contrastive} with a ResNet-based generator of 9 residual blocks ($G_{S \rightarrow T}$) and a PatchGAN discriminator ($D_T$).

\subsubsection{Training}
When training the CUT-based domain adaptation network $G_{S \rightarrow T}$, the image is randomly cropped to 320 $\times$ 320 pixels and the network is trained for 6 epochs with learning rate 2.0 $\times$ $10^{-3}$, batch size 10 and Adam optimizer. The weights in equation \ref{eq5} are set to $\lambda_{GAN}$ = 1, $\lambda_{S}$ = 1 and $\lambda_{Y}$ = 1. In order to compute the multi-layer PatchNCE loss ($\Loss_{PatchNCE}$), features are extracted from 5 layers ($L$ = 5), which correspond to RGB pixels, the first and second downsampling convolution, and the first and the fifth residual block. These layers correspond to receptive fields of sizes (i.e. patch sizes) 1$\times$1, 9$\times$9, 15$\times$15, 35$\times$35, and 99$\times$99. For features of each layer, 256 ($S$ = 256) random spatial locations are sampled, and a 2-layer MLP $H_l$ is employed to extract 256-dim final features. To train the final SCR network $f_T$, the image is also randomly cropped to 320 $\times$ 320 pixels and the network is trained using the Adam optimizer with learning rate 1.0 $\times$ $10^{-4}$, weight decay 1.0 $\times$ $10^{-5}$, and batch size 48. Because our \MethodName\ pipeline can generate innumerate training data, we do not perform other data augmentation methods besides random cropping. All the hyper-parameters and the stopping criterion are selected based on the experimental results of an independent validation dataset.

\section{DATASET GENERATION}

We evaluate the pose estimation performance of several techniques within our laboratory space (5.8 meters wide, 14.3 meters long, 3.0 meters high).
We capture or generate all necessary datasets for training, validation, and testing to ensure a fair comparison among the techniques.

\subsection{Laser scan}
\GS{While there is nothing specific to point clouds about \MethodName, for simplicity we use a Leica BLK360 laser scanner to capture a color point cloud of our robotics lab space. To reduce occlusions, we merge 16 scans into a single point cloud of 118M points,  %
each storing location (X, Y, Z) and color (R, G, B) information.}

\subsection{Synthetic Images} %
Synthetic images with corresponding scene coordinate labels are generated by placing a virtual camera with known random pose in the space and projecting the point cloud onto the virtual image plane using intrinsic parameters measured from the device camera. We render 100k synthetic images with corresponding scene coordinate labels from virtual camera poses drawn from a similar distribution as the device images (described below). These images and labels can be considered a representative sample set from the source distribution $X_S$ and $Y_S$. 
The lab is equipped with a WorldViz infrared tracking system that serves as the reference coordinate frame and provides poses for evaluation purposes (this is the expensive step that DANCE seeks to circumvent). The transformation between WorldViz and the point cloud(s) is established by recording the coordinates of fixed WorldViz markers with respect to each frame and aligning them via the iterative closest point algorithm.

\subsection{Camera Images}
We sample the target distribution $X_T$ (capture real photos) by moving an iPhone 6 in the space. The phone is mounted horizontally on a wheeled cart which is pushed manually through the space in order to mimic a robot with a fixed RGB camera. The ground truth pose of the camera is determined by tracking multiple WorldViz markers placed on the cart and by establishing the transformation between the cart and camera frames. Camera images are extracted from videos captured over four trajectories and downsampled to $640\times360$ pixels. 
We dedicate two trajectories (28411 images) for training the baseline networks for comparison with DANCE, one trajectory (1637 images) for validation, and one trajectory (2104 images) for testing purposes. The WorldViz ground truth pose labels are collected for all the photos, but these labels are not needed for the training of the DANCE pipeline. These pose labels are only used for training the fully-supervised baselines and for evaluation. %

\subsection{Domain Gap}
\GS{
Although we do not need to know where the domain gap stems from in order to bridge it, we hypothesize that in our case it stems from the simplicity of the rendering method. An image rendered from a point cloud has inevitable holes due to occlusions and splatting artefacts, so the resulting synthetic image largely differs from a photo of the scene.
}

\begin{figure*}[th]
\centering
  \vspace{0.1cm}
  \includegraphics[width=\textwidth]{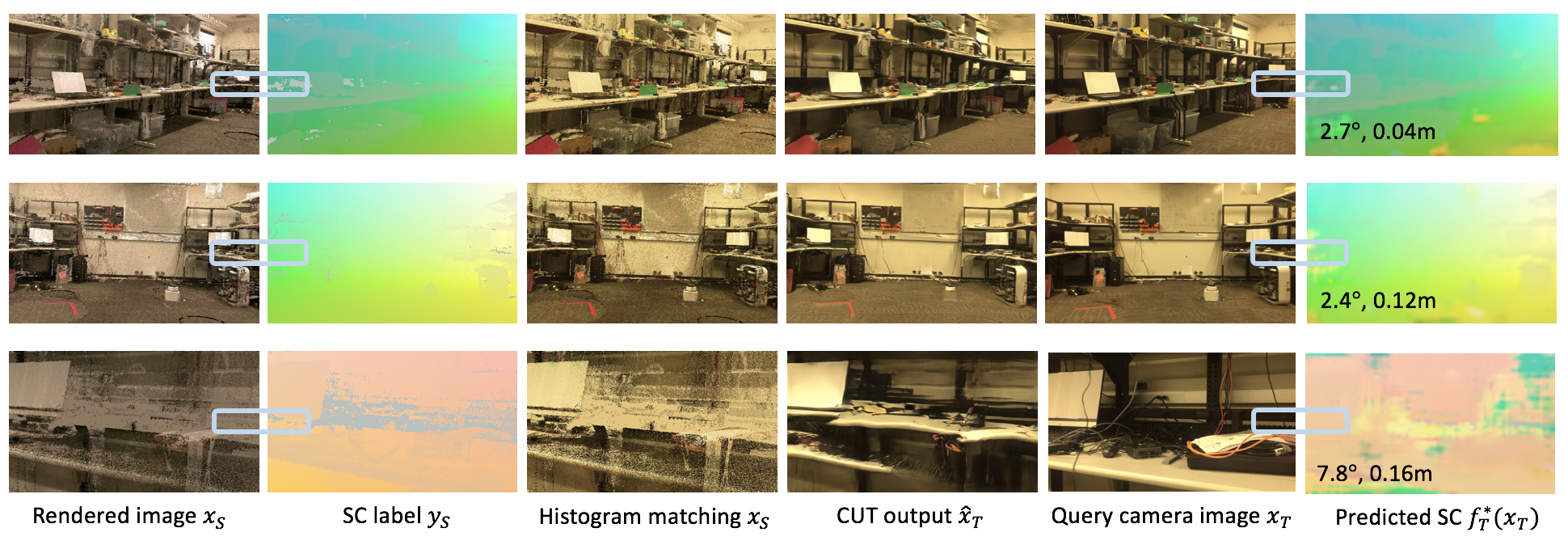}
  \vspace{-0.1cm}
  \caption{Three samples are shown with some intermediate results. The numbers on the predicted scene coordinates (SC) indicate estimated pose errors. The last row is an unsatisfactory case because the query image has small field of view and the corresponding location is poorly covered by the point cloud. For illustrative purposes, the rendered (source) images $X_S$ and camera (target) images $X_T$ are shown in the same poses. In practice, $X_S$ and $X_T$ do not have any pairwise relationship.}
  \label{5Images}
  \vspace{-0.5cm}
\end{figure*}

\section{EXPERIMENTAL RESULTS}

We evaluate the predicted pose error of each technique with respect to the measured ground truth (WorldViz) pose. All the results are reported on the 2104 test images.

\GS{\paragraph{Comparing training strategies} We compare side by side multiple training strategies of the same SCR network, including the original, expensive fully supervised labels, and we show that our proposed strategy can achieve comparable median performance.} As summarized in Table \ref{table:1}, we first quantify an intuitive lower bound (a) and upper bound (e) of the $\MethodName$ pipeline where the SCR networks are trained with different strategies. 
When the SCR network is trained on the 100k synthetic images with corresponding scene coordinate ground truth (without any domain adaptation, ($X_S$, $Y_S$)) and tested with real camera images $X_T$, it fails (Table \ref{table:1}(a)). This lower bound (Blind Transfer) performance indicates that the domain gap between the rendered and camera images is significant.
When the SCR network is trained on the 28411 real images with scene coordinate labels (rendered from the point cloud using ground truth poses only for evaluation), $2.9^{\circ}$, $0.17\m$ median error is obtained (Table \ref{table:1}(e)). This indicates that if there is no domain gap between the training and testing images, our proposed SCR and PnP-RANSAC pipeline can achieve good performance.

\definecolor{dgreen}{RGB}{0, 180, 0}
\definecolor{dyellow}{RGB}{180, 180, 0}
\definecolor{dred}{RGB}{180, 0, 0}
\definecolor{dorange}{RGB}{255, 140, 18}

\begin{table}[htb!]
\begin{tabular}{| p{22.0mm} | p{16.7mm} | p{16.5mm} | p{11.0mm} |} 
 \hline

 Method &    Median error &    95\%-tile error &    Requires \\
 \hline
 \hline
 
  PoseNet~\cite{Kendall2015PoseNet} &    $4.2^{\circ}, 0.28\m$ &    $16.9^{\circ}, 0.69\m$ &   \textcolor{dorange}{RP} \\ %
 \hline
 UcoSLAM~\cite{MunozSalinas2020UcoSLAM} &    $2.7^{\circ}, 0.08\m$ &  9.7\% invalid &    \textcolor{dgreen}{R+}\\ %
 \hline
  \hline
 
 (a): Blind Transfer %
 &    $110^{\circ}, 7.57\m$ &    $175^{\circ}, 16.5\m$ &    \textcolor{dgreen}{SSC} \\    %
 \hline
 (b): (a) + Hist. match. &    $14.1^{\circ}, 1.09\m$ &    $154^{\circ}, 15.3\m$ &  \textcolor{dgreen}{SSC},\textcolor{dgreen}{R} \\    %
 \hline
(c): (b) + CycleGAN &    $4.2^{\circ}, 0.22\m$ &    $96.5^{\circ}, 6.58\m$ &    \textcolor{dgreen}{SSC},\textcolor{dgreen}{R} \\    %
 \hline
 (d): (b) + CUT (\bf{\MethodName})  & $3.0^{\circ}, 0.14\,m$ & $25.7^{\circ}, 0.95\m$ & \textcolor{dgreen}{SSC},\textcolor{dgreen}{R} \\    %
 \hline
 (e): Fully Supervised~\cite{Brachmann2020DSACStar} %
  &    $2.9^{\circ}, 0.17\m$ &    $11.1^{\circ}, 0.47\m$ &    \textcolor{dred}{RSC} \\    %
\hline
\end{tabular}
\caption{Comparison of existing pose estimation methods and variants of our \MethodName\, proposal. 
Requirements: 
\textcolor{dgreen}{R} Real images (lowest cost); %
\textcolor{dgreen}{R+} Real image sequence (low);
\textcolor{dgreen}{SSC} Synthetic images with Scene Coordinates (low);
\textcolor{dorange}{RP} Real images with Poses (high); or \textcolor{dred}{RSC} Scene Coordinates (highest). %
The synthetic labeled images are rendered from a color point cloud of the same scene. Compared to PoseNet and its fully supervised counterpart (e), \MethodName\ achieves similar performance with significantly lower data acquisition cost. Compared to UcoSLAM which has $9.7\%$ invalid pose estimates, \MethodName\, has better tail performance and does not require the real images to be in sequence. (e) is a componentwise training variant of DSAC~\cite{Brachmann2020DSACStar}. All the baseline methods are retrained using our training data.}
\label{table:1}
\end{table}

When we trained the SCR network within the proposed unsupervised deep domain adaptation framework (100k domain adapted labeled synthetic images ($G^*_{S \rightarrow T}(X_S)$, $Y_S$)), $\MethodName$ (Table \ref{table:1}(d)) outperforms the lower bound method (no domain adaptation, Table \ref{table:1}(a)) by a large margin and is on par with the upper bound method (full supervision) in terms of median error (Table \ref{table:1}(e)). We do not claim 0.14m median error of \MethodName\ is better than 0.17m of the upper bound method due to the lack of statistical test (Table \ref{table:1} (d) vs (e)). This indicates that the domain gap necessitates the application of domain adaptation techniques, and that our proposed training pipeline is effective at narrowing the domain gap between the rendered images $X_S$ and real camera images $X_T$. It is worth noting that the fully supervised upper bound method (Table \ref{table:1}(e)) is a componentwise training variant of DSAC~\cite{Brachmann2020DSACStar} (a SCR based framework) where the SCR network and PnP-RANSAC were trained in an end-to-end manner. DSAC~\cite{Brachmann2020DSACStar} showed that end-to-end training can only provide marginal performance gain compared with componentwise training, which indicates the upper bound method is a strong baseline for comparison.

Compared with other existing methods, $\MethodName$ achieves lower median errors than the fully supervised PoseNet~\cite{Kendall2015PoseNet} and comparable median errors to a fully supervised SCR based method~\cite{Brachmann2020DSACStar} (Table \ref{table:1}(e)) with much lower deployment overhead. Specifically, PoseNet requires camera images with ground truth poses for training (28411 real images with ground truth poses in our experiments) while our pipeline only requires rendered images with ground truth scene coordinates (100k ($X_S$, $Y_S$)) and unlabeled camera images (28411 real images $X_T$ without labels). The unlabeled camera images are only used to provide target domain information to train the mapping network $G_{S \rightarrow T}$ in our pipeline. Though PoseNet is not the state-of-the-art learning based method, the results demonstrate that our unsupervised pipeline is able to achieve better localization performance than a fully supervised approach. Furthermore, using \MethodName\ to train an SCR network achieves comparable median performance to training the same SCR network using the harder to acquire fully supervised labels (Table \ref{table:1}(e)) thereby demonstrating that the \MethodName\ pipeline can bridge the domain gap sufficiently to allow comparable performance to fully supervised methods.

\begin{figure}[!ht]
\hspace{-0.1cm}
\includegraphics[width=1.0\columnwidth]{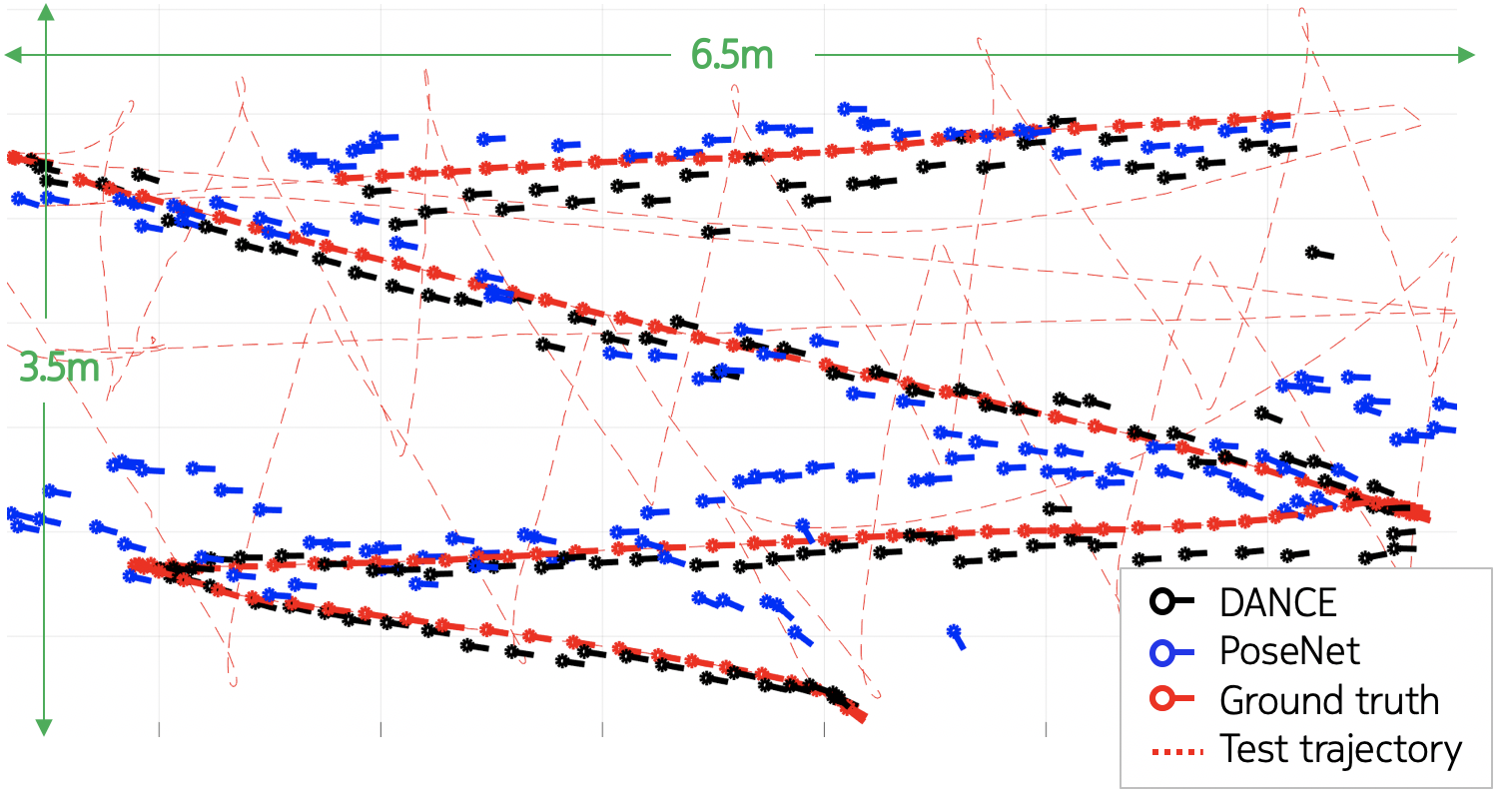}
\caption{
Top-down view of the room. The full test trajectory is shown in the background (dash red line), and pose estimates (solid arrows) on a subset of the trajectory are shown in the foreground for qualitative comparison. The solid arrows indicate the camera location and orientation. The statistics in Table~\ref{table:1} are w.r.t. the full test trajectory.
} %
\label{poseEstPlan}
\vspace{-0.25cm}
\end{figure}

\GS{In summary, as shown in Table~\ref{table:1} and Figure~\ref{poseEstPlan}, 
the \MethodName-trained SCR network can achieve performance comparable to fully supervised PoseNet, but without the need for tedious real pose labels. 
}

\paragraph{Comparison with feature-based relocalization}
We also compare the relocalization performance of our method with a state-of-the-art, feature-based SLAM system called UcoSLAM~\cite{MunozSalinas2020UcoSLAM} (Table \ref{table:1}). This is a recent variant of the popular ORB-SLAM2~\cite{MurArtal2017ORBSLAM2} with support for fiducial landmarks, map loading/saving, and a highly speed-optimized version of the DBoW2 \cite{GalvezLopez2012DBoW} bag-of-words image matcher. 
We build the SLAM map with the same real image sequences we use in the training of our proposed method (two sequences with 28411 camera images). At test time, we enforce relocalization for each frame of the test sequence. We acknowledge this is not the ideal use case for a SLAM algorithm, but it makes a fair comparison with a single-frame localization method possible.
While \MethodName\, performs slightly worse than UcoSLAM in terms of median errors, it is much more robust as it returns a valid pose for every frame while UcoSLAM sometimes fails to get valid results ($25.7^{\circ}$, 0.95m 95\%-tile error vs 205 frames out of 2104 testing frames are invalid, Table \ref{table:1}). It is also worth noting that SLAM requires a whole image sequence to build a 3D map while our proposed pipeline only requires unordered images to provide target domain information for training $G_{S \rightarrow T}$.

\paragraph{Other domain adaptation methods}
To investigate the efficacy of each domain adaptation component in our proposed pipeline (\MethodName), we compare various architectural options in Table \ref{table:1}. The lower bound (a) error is $110^{\circ}$, $7.57\m$ when training the SCR network on the synthetic images and testing on the real images. We then perform histogram matching (b) from rendered images to camera images and train the SCR network on these transformed images. Histogram matching improves the median error to $14.1^{\circ}$, $1.09\m$.
Next, the image level adaptation network $G_{S \rightarrow T}$ is trained with different GAN frameworks (CycleGAN~\cite{Zhu2017CycleGAN} vs CUT~\cite{park2020contrastive}) to map the rendered images after histogram matching to camera images (Table \ref{table:1} (c) vs (d)). Our \MethodName\ pipeline adopts the CUT framework to train the adaptation network $G_{S \rightarrow T}$, which was shown to have better unpaired image-to-image translation power than CycleGAN. By training the SCR network on these domain adapted labeled rendered images ($G^*_{S \rightarrow T}(X_S)$, $Y_S$), \MethodName\-CUT (Table \ref{table:1}(d)) outperforms DANCE-CycleGAN (Table \ref{table:1}(c)) by a significant margin. This indicates that a better unpaired image-to-image translation GAN model can further improve our \MethodName\ pipeline.

\paragraph{Other 3D representations}
\GS{
Besides a color point cloud (the 3D representation in DANCE pipeline) captured from a laser scan, we also tested our method in case the 3D representation is a SfM model of the space. While sparse SfM point clouds can be used for feature-based localization, generating scene coordinates requires a dense model. We performed dense reconstruction from our training images using Colmap~\footnote{\url{https://colmap.github.io/}}, but the quality of the resulting 3D representation was poor with significant distortion and missing areas. We concluded that an SfM pipeline is not necessarily suitable for building the 3D representation of the scene in order to generate the labeled rendered images. In the future, better domain adaptation methods might be able to bridge such even larger domain gap. It is an interesting question what is the minimally required quality of a reconstruction for our domain adaptation technique to work, we leave this analysis for future work.
}

\paragraph{Other coordinate regression networks}
\GS{Note that our primary goal was to simplify the training process of pose (or scene coordinate) regression networks in general, in order to make this family of methods more accessible, and chose the fully supervised PoseNet as one of well-known baselines for comparison. We have shown to achieve results similar to these fully supervised methods but only at a fraction of the cost.
}
Since its original publication, several methods have improved on PoseNet, and we anticipate that swapping to a more powerful SCR method may improve performance. This is indeed possible in our general training framework and we see this as a key strength of our framework. There is no assumption on the pose estimation network (PoseNet or other), there is no assumption on the input 3D representation (point cloud), there is no strict assumption on the domain adaptation method used (we tested a CycleGAN-like pipeline as well as CUT). Furthermore, while our general training framework could be applied to other, newer scene coordinate regression methods, we also expect that with better laser scanners, let alone better domain adaptation methods in the future, the accuracy could be even further improved.

\section{CONCLUSIONS}
We have shown that it is possible to train a neural network to perform the task of scene coordinate regression for monocular camera pose estimation on real images using only synthetic labeled images and a pool of unordered unlabeled photos. Our proposal achieves performance comparable with existing fully supervised techniques but with significantly lower overhead cost. 
These existing techniques require photos with ground truth camera pose labels, which are typically obtained using cumbersome motion capture systems that track markers mounted on the camera. For each room or environment, the motion capture system would need to be deployed to generate a new set of labeled training photos.   

In contrast to existing approaches, deploying \MethodName\ in a new room is simpler, requiring (unlabeled) images along with a dense 3D representation of the room to generate synthetic labeled images. Our dense representation was captured with the push of a button using a tripod-mounted Leica BLK360 scanner. Alternatively, one could potentially use even cheaper capture systems such as recent iOS devices which come equipped with LiDAR. In general, we believe our \MethodName\ pipeline will continue to benefit from both the rapid development of 3D capture techniques and more powerful unpaired image-to-image translation models. 

Furthermore, there is no reason why this pipeline cannot be applied in tasks beyond pose estimation, and modalities beyond images. Any task for which there is an abundance of unlabeled samples, and for which the construction of a crude simulation is easier than the direct gathering of labeled data should be amenable to this technique. %

\section*{ACKNOWLEDGMENTS}
We sincerely thank Kriszti\'an Varga, Wei Zhu, and Prof. Chen Feng for our fruitful discussions as well as Spencer Langerman for the elucidating edits.

{\small
\bibliographystyle{ieee}
\bibliography{dance}
}

\end{document}